# Steady State Resource Allocation Analysis of the Stochastic Diffusion Search

Slawomir J. Nasuto[1] and Mark J. Bishop,

Department of Cybernetics, The University of Reading, Whiteknights RG6 2AY, Reading, UK, [1]sjn@cyber.rdg.ac.uk

*Abstract.*

This article presents the long-term behaviour analysis of Stochastic Diffusion Search (SDS), a distributed agent-based system for best-fit pattern matching. SDS operates by allocating simple agents into different regions of the search space. Agents independently pose hypotheses about the presence of the pattern in the search space and its potential distortion. Assuming a compositional structure of hypotheses about pattern matching agents perform an inference on the basis of partial evidence from the hypothesised solution. Agents posing mutually consistent hypotheses about the pattern support each other and inhibit agents with inconsistent hypotheses. This results in the emergence of a stable agent population identifying the desired solution. Positive feedback via diffusion of information between the agents significantly contributes to the speed with which the solution population is formed.

The formulation of the SDS model in terms of interacting Markov Chains enables its characterisation in terms of the allocation of agents, or computational resources. The analysis characterises the stationary probability distribution of the activity of agents, which leads to the characterisation of the solution population in terms of its similarity to the target pattern.

**Keywords**: generalised Ehrenfest Urn model, interacting Markov Chains, nonstationary processes, resource allocation, best-fit search, distributed agents based computation,

## 1. Introduction.

In this paper we investigate the steady state behaviour of a probabilistic heuristic search method called Stochastic Diffusion Search (SDS), [Bishop, 1989]. SDS is a generic, distributed, agent-based method for solving the best-fit matching problem. Many fundamental problems in Computer Science, Artificial Intelligence or Bioinformatics may be formulated in terms of pattern matching or search. Examples abound in DNA or protein structure prediction, where virtually all the deterministic methods employed are solving variations of the string matching problem [Gusfield, 1997]. Other areas of application include information retrieval, wireless communications, speech recognition, or signal processing. The classical exact string matching problem has been extended to the approximate matching, where one allows for a pre-specified number of errors to occur. Other variations include consideration of various distances used for the determination of similarity between patterns, see [Navarro, 1998] for an extensive overview of an approximate string matching algorithms. String matching can be generalised to tree matching and many algorithms used for string matching are easily adapted to this significant class of problems [van Leeuven, 1990]. Thus, it is very important to develop new efficient methods of solving string-matching problems.

The last decades have witnessed an increased interest in various forms of heuristic search methods, as alternative solutions to hard computational problems. This arose due to recognition of the limitations of fully specified deterministic methods based on sequential operation and computation. Such heuristic methods include Genetic Algorithms, Evolutionary Strategies, Ant Colony, or Simulated Annealing, [Holland, 1975; Back, 1996; Colorni et al, 1991; Kirkpatrick et al, 1983]. They often base their computation on some form of iterative parallel random sampling of the search space by a population of computational agents, where the sampling bias is a function of coupling mechanisms between the agents and thus is specific to the particular method employed. For example Genetic Algorithms [Goldberg, 1989] are loosely inspired by the natural evolution and the mechanisms by which GA agents sample the search space are most often described in terms of gene recombination and mutation. Ant Colony Optimisation is based on the modelling of the interaction between simple organisms like ants or termites [Dorigo, 1992]. These methods can be collectively described as heuristic search methods, as their generic formulation is based on very simplified intuitions about some natural processes. However, in spite of considerable interest in these algorithms and their wide areas of applications, they still lack a standard formal framework enabling principled inference about their properties. It can be argued that the very central and indeed appealing feature of these heuristic search methods (i.e. imitating solutions to hard computational problems found by Nature) impedes development of their theoretical basis.

However, randomness in computation has also been employed in more classical computational schemes, like RANSAC, Random Global Optimisation, Monte Carlo Markov Chains or related to them Particle Filters [Fischler, Bolles, 1981; Zhigljavsky, 1991; Gilks et al, 1996; Doucet et al, 2000]. In these algorithms biased random sampling is employed in order to avoid local minima over the search spaces or to approximate values of otherwise intractable functions. The theoretical framework of these algorithms is usually within the area of discrete stochastic processes and the understanding of their properties and behaviour is much more sound than that of their heuristic counterparts.

Simulated Annealing and Extremal Optimisation belong to a small class of heuristic search algorithms taking inspiration from physical processes, which nevertheless enjoy a sound theoretical basis [van Laarhoven, Aarts, 1987; Bak et al, 1987].

In this article we describe steady state analysis of Stochastic Diffusion Search (SDS) [Bishop, 1989] an efficient heuristic search method based on distributed computation.

SDS operates by allocating simple agents into different regions of the problem search space. Agents independently pose hypotheses about the presence of a target pattern and its potential distortion in the search space. Agents utilise the compositional structure of hypotheses about matching patterns by performing an inference on the basis of only partial evidence. A population of agents posing mutually consistent hypotheses about the pattern signifies the desired solution. This population is formed very rapidly as the search progresses due to diffusion of information between the agents. SDS has been successfully applied in various challenging engineering problems, e.g. invariant facial features location [Bishop, Torr, 1992], real time lip tracking [Grech-Cini, 1995] and in the navigational system of an autonomous vehicle [Beattie, Bishop, 1998].

Some fundamental properties of SDS have been previously investigated within the framework of Markov Chains theory [Nasuto, Bishop, 1999; Nasuto et al, 1998]. In particular, it has been proven that SDS converges to the globally best fit matching solution to a given template [Nasuto, Bishop, 1999] and that its time complexity is sub-linear in the search space size [Nasuto et al, 1998], thus placing it among the state of the art search algorithms [van Leeuven, 1990].

In this paper we investigate how SDS achieves its tasks by distributing computational resources while focusing on the solution. By complementing previous results on global convergence and time complexity of SDS, the resource allocation analysis constitutes a significant advancement of our understanding of this algorithm operation. It appears that SDS is very robust and is capable of differentiating between very small differences between patterns. This is achieved due to positive feedback leading to a rapid growth of clusters inducing strong competition between them due to the limitation of agent resources. Thus SDS converges to the best-fit solution by allocating to it the largest cluster of agents. However, the dynamic aspect of the process ascertains that the remaining amount of agents, currently not assigned to any cluster, continue to explore the rest of the search space, thus ensuring the global convergence. In this paper we characterised quantitatively the distribution of agents during the search and obtained the expected size and variance of the cluster corresponding to the desired solution. We also characterised their dependence on the characteristics of the search space and the quality of the best solution.

The model of SDS used in this paper is based on a novel concept of interacting Markov Chains [Nasuto, 1999]. It concentrates on modelling the behaviour of individual agents as opposed to modelling the whole population. This is a clear advantage as the size of the state space corresponding to the whole population of agents grows quadratically with their number; the state space of single agents is very small. However, in SDS agents are interacting with each other and their actions depend on actions of other agents. The evolution of the whole population of agents is thus reflected in coupling between the Markov Chains. The dependence of the transition probabilities of single chains not only on their immediate past but also on the state of other chains introduces nonlinearity into the system description. However, it is possible to show that there exists a system of decoupled chains such that its long-term behaviour is the same as that of interacting Markov Chains modelling SDS. The decoupled system of Markov Chains with such behaviour forms a generalisation of an Ehrenfest Urn model of irreversible processes introduced in statistical physics at the beginning of this century [Ehrenfest, Ehrenfest, 1907]. The formulation of an Ehrenfest Urn model with an equivalent long-term behaviour constitutes a basis for the quantitative characterisation of SDS steady state distribution of the computational resources discussed in this paper.

The paper is organised as follows. Section 2 introduces Stochastic Diffusion Search and presents the relationship between its model based on interacting Markov Chains and the generalised Ehrenfest Urn model. Section 3 characterises the long-term behaviour of an agent in SDS. This, together with characterisation of the dependence of the steady state agent distribution on the search conditions, presented in the next section, forms the basis of the characterisation of resource allocation. The numerical simulations of SDS steady state behaviour are presented in section 5. The final section discusses the results and future work.

*2. Stochastic Diffusion Search and the Ehrenfest Urn Model.*

Stochastic Diffusion Search employs a population of simple computational units or agents searching in parallel for the pre-specified template in the search space. Agents take advantage of the compositional nature of the statistical inference about templates. This property ensures that the hypothesis about the template can be factored into hypotheses about its constituents. At each step agents perform an inference about the presence of the template in the currently investigated region of the search space by posing and testing hypothesis about a randomly selected template's sub-component. Each of the agents makes a decision about which region of the search space to explore on the basis of its history and the information it gets from other agents in the so called diffusion phase, see Figure 1. This stage enables agents that failed their inferential step (*inactive agents*) to bias their consecutive sampling of the search space towards the region of the search space, which led to a successful test by another, randomly chosen agent (*active agent*). This mechanism ensures that regions with higher probability of positive testing will gradually attract higher numbers of agents; regions of low positive testing attracting agents only sporadically. Alternatively, inactive agents have a possibility to resample entirely new positions from the search space, had there been no sufficient reason to copy the position inspected by the chosen agent. While in the next iteration active agents attend to the same position they tested previously thus exploiting current patterns for further evidence regarding matching to the template. The random resampling of the search space by inactive agents ensures that it will be sufficiently explored.

Patterns of highest overlap with the template in the search space induce a form of local positive feedback because the more agents investigate them the more likely is increase of their number in the next step due to the diffusion phase. This local positive feedback induces a competition between best-fit patterns due to limited (computational) resources; the largest cluster of agents concentrated on the globally best solution emerges rapidly suppressing sub-optimal solutions. Thus, SDS performs the *best-fit* pattern matching; it will find the pattern with the highest overlap with the template. This is an extension of a concept of exact and approximate matching in computer science in which the problem is to locate either *exact copy* or an instantiation differing from the template by a pre-specified number of components [van Leeuwen, 1990].

```
       INITIALISE;            #agents are initialised randomly
       REPEAT;                #for all agents in the population
           DIFFUSION          #information by communication
           TEST               #partial hypotheses testing by agents
       UNTIL TERMINATION;
```

**Figure 1.** Schematic of Stochastic Diffusion Search

*Interacting Markov Chains Model of the Stochastic Diffusion Search.*
Consider a noiseless search space in which there exists a unique object with a non-zero overlap with the template - a desired solution. Assume that upon a random choice of a feature for testing a hypothesis about the solution location, an agent may fail to recognise the best solution with a probability $p^- > 0$. Assume further that in the $n^{th}$ iteration $m$ out of a total of N agents are *active* (i.e. tested successfully). Then the following transition probability matrix describes a one-step evolution of an agent:

$$P_n = \begin{array}{c} a \\ n \end{array} \begin{bmatrix} 1 - p^- & p^- \\ p_1^n & 1 - p_1^n \end{bmatrix},$$

where

$$p_1^n = \tfrac{m}{N}(1 - p^-) + (1 - \tfrac{m}{N}) p_m (1 - p^-),$$

$p_m$ is a probability of locating the solution in the search space by uniformly random sampling and the active agent pointing to the correct solution is denoted by '*a*' and an *inactive* agent (i.e. agent that failed the test) by '*n*'.

Thus, if an agent was active in the previous iteration then it will continue to test the pattern at the same position but using another, randomly chosen sub-component. Therefore with probability $1-p^-$ it may remain active (tests positive), otherwise it becomes inactive. An inactive agent may become active either by choosing at random an active agent and testing positively at its position (the first term in $p_1^n$) or otherwise by testing positively upon a random resampling of the search space.

It is apparent that the above stochastic process modelling an evolution of an agent is a nonhomogenous Markov Chain. Nonhomogeneity stems from the fact that the entries of the second row of the probability transition matrix $P$ are not constant in time but change as search progresses thus making the matrix $P_n$ time dependent. This is because the probability of an inactive agent becoming active in the next iteration depends on the normalised number of active agents in the previous iteration. The time dependence of P reflects the interactions with other agents in the population. Thus, the population of Markov Chains defined above describes the evolution of a population of agents in SDS.

The model capturing the agent's behaviour is similar to the model formulated by Ehrenfests in 1907 in order to resolve an apparent irreversibility paradox in the context of statistical mechanics [Ehrenfest, Ehrenfest, 1907]. In fact, we will demonstrate that the long-term behaviour of SDS can be recovered from an appropriately constructed generalised Ehrenfest Urn model.

The Ehrenfest Urn model consists of two urns containing in total *N* balls. At the beginning there are *k* balls in the first urn and *N-k* in the other. A ball is chosen at random from a uniform distribution over all *N* balls and is placed in the other urn. The process relaxes to the stable state. In this state it remains most of the time in a quasi-equilibrium, which corresponds to approximately equal numbers of balls in both urns, subject to small fluctuations. Whittle discussed a generalisation of this model consisting of a collection of N simultaneous and statistically independent 2-state Markov chains governed by the same transition matrix, [Whittle, 1986],

$$\begin{bmatrix} f_{11} & f_{12} \\ f_{21} & f_{22} \end{bmatrix},$$

Thus a ball moving between the urns corresponds to a single Markov Chain. Its transition probabilities between the two urns, $f_{12}$ and $f_{21}$, are in general not equal.

There is an important difference between the generalised Ehrenfest Urn Model and Interacting Markov Chains model of SDS. In the former the Markov Chains are independent and stationary, whereas in the latter the Markov Chains are interacting. In addition, the coupling between Markov Chains induces their nonhomogeneity - their transition probability matrices change in time. However, we will prove that the long-term behaviour of the Interactive Markov Chains model can be described in terms of an appropriately constructed generalised Ehrenfest Urn model.

The analysis based on this model will characterise the stationary probability distribution of the activity of a population of agents. This will allow calculation of the expected number of agents forming a cluster corresponding to the best-fit solution as well as the variation of the cluster size. The model will also enable characterisation of the long-term behaviour of SDS in terms of the statistical properties of the search space.

*3. Long term behaviour of the Stochastic Diffusion Search.*

In order to investigate the Markov Chain model of agent's evolution we will establish first its long-term behaviour. The following propositions about single agent evolution can be proven:

**Proposition 1.** The sequence $\{P_n\}$ of stochastic matrices describing the evolution of an agent in the Stochastic Diffusion Search is weakly ergodic.

*Proof.*     See Appendix.

**Proposition 2.** $\{P_n\}$ is asymptotically stationary.

*Proof.*     See Appendix.

**Proposition 3.** The sequence $\{P_n\}$ is strongly ergodic.

*Proof.* Strong ergodicity of $\{P_n\}$ follows as a consequence of the above propositions, and theorem 4.11 in [Seneta, 1973].

From Propositions 1-3 it follows that the Markov Chain corresponding to the evolution of a single agent is asymptotically homogenous, i.e. the limit $P = \lim_{i \to \infty} P_i$ exists. Thus this process behaves, as it approaches the equilibrium, more and more like a homogenous Markov Chain with the transition probability matrix $P$. Therefore, instead of considering a population of interacting Markov Chains we will construct and consider a generalised Ehrenfest Urn model consisting of homogenous Markov Chains with the transition probability matrix $P$.

First we will obtain an explicit formula for the matrix $P$. The strong ergodicity of a nonhomogenous Markov Chain amounts to the existence of its (unique) equilibrium probability distribution. In other words, a stochastic process describing the time evolution of an agent as it visits states of its state space with frequencies, which can be characterised via the limiting probability distribution.

Thus, in order to find the limit probability distribution one has to find the solution of the fixed-point equation

$$S(p, P) = (p, P)$$

(see Appendix for the definition of the mapping $S$).

This amounts to solving a system of two equations

$$\begin{cases} \pi P = \pi, \\ p_1 = (1 - p^-)\pi_1 + (1 - p^-)(1 - \pi_1)p_m \end{cases} \quad (1)$$

The first equation is an equation for an eigenvalue 1 of a two-by-two stochastic matrix $P$, for which the eigenvector has the form

$$\pi = (\pi_1, 1 - \pi_1) = \left(\frac{p_1}{p_1 + p^-}, \frac{p^-}{p_1 + p^-}\right) \quad (2)$$

This makes it possible to find the solution in the special case when $p_m=0$ (no solution in the search space). It follows, that the initial distribution of agents is concentrated entirely on the inactive state and from the latter it follows that $p_1=0$ so, as expected,

$$\pi = (0, 1)$$

i.e. an agent will always remain inactive.

To find the solution in the general case, assume that $p_m>0$. From equation (2) and the second equation of (1) it follows that

$$\pi_1\left[(1 - p^-)\pi_1 + (1 - p^-)(1 - \pi_1)p_m + p^-\right] = (1 - p^-)\pi_1 + (1 - p^-)(1 - \pi_1)p_m,$$

which after rearrangement leads to a quadratic equation in $\pi_1$

$$(1 - p^-)(1 - p_m)\pi_1^2 + \left[2(1 - p^-)p_m + 2p^- - 1\right]\pi_1 - (1 - p^-)p_m = 0$$

This can be written in the form

$$(1 - p^-)(1 - p_m)\pi_1^2 - \left[2(1 - p^-)(1 - p_m) - 1\right]\pi_1 - (1 - p^-)p_m = 0$$

This equation has two solutions because the condition

$$\left[2(1 - p^-)(1 - p_m) - 1\right]^2 + 4(1 - p^-)^2(1 - p_m)p_m \geq 0$$

is always fulfilled. These solutions are as follows

$$\pi_i = \frac{2(1 - p^-)(1 - p_m) - 1 \pm \sqrt{\left[2(1 - p^-)(1 - p_m) - 1\right]^2 + 4(1 - p^-)^2(1 - p_m)p_m}}{2(1 - p^-)(1 - p_m)}, \quad i = 1, 2$$

Straightforward analysis of the above solutions implies that only one of them can be regarded as a solution to the problem. Namely, the desired equilibrium probability distribution is

$$\pi = (\pi_1, \pi_2) =$$
$$\left( \frac{2(1-p^-)(1-p_m) - 1 + \sqrt{[2(1-p^-)(1-p_m) - 1]^2 + 4(1-p^-)^2(1-p_m)p_m}}{2(1-p^-)(1-p_m)}, \right. \quad (3)$$
$$\left. \frac{1 - \sqrt{[2(1-p^-)(1-p_m) - 1]^2 + 4(1-p^-)^2(1-p_m)p_m}}{2(1-p^-)(1-p_m)} \right)$$

As the long term evolution of an agent is approaching the evolution of a homogenous Markov Chain with the transition matrix $P$, we can characterise the limit behaviour of the Interactive Markov Chains model of SDS by finding the behaviour for the corresponding generalised Ehrenfest Urn model, in which the probability transition matrix $P$ governing the transitions of a ball is

$$P = \begin{matrix} & a & n \\ a & \\ n \end{matrix} \begin{bmatrix} 1-p^- & p^- \\ p_1 & 1-p_1 \end{bmatrix}$$

where

$$p_1 = (1-p^-)\pi_1 + (1-p^-)(1-\pi_1)p_m$$

and $\pi_1$ is given by (3).

It is important to note that in the above Ehrenfest Urn model the transition of balls between the urns is now mutually independent. This is because the dependence between Markov Chains in the Interacting Markov Chains model of SDS was reflected in the changes of their probability transition matrices. By using a limit probability transition matrix $P$ we make the balls independent. Thus in order to characterise the limit probability distribution we will proceed analogously to [Whittle, 1986].

Modelling SDS via evolution of single agents, as proposed in the model, implies distinguishability of agents. This is true also in the case of the balls in the corresponding generalised Ehrenfest Urn model. From this it follows that the state space of the ensemble of Markov Chains corresponding to $N$ balls in the constructed Ehrenfest Urn model consists of $N$-tuples $x = (x_1, ..., x_N)$ where the $i^{th}$ component takes on a value 1 or 0 depending on whether the $i^{th}$ ball (agent in SDS) was in urn 1 or 2 (was active or not). Because of ergodicity of Markov Chains describing evolution of balls and their independence there will be a unique asymptotic limit probability distribution given by:

$$\Pi(x) = \pi_1^{a[x]} \pi_2^{N-a[x]}, \quad (4)$$

where $a[x]$ denotes number of balls in the urn 1 (active agents) corresponding to the state $x$ and $(\pi_1, \pi_2)$ is the two state asymptotic distribution given by (3).

In order to establish the equilibrium probability distribution of SDS one has to abandon the distinguishability of balls implied by the construction of the generalised Ehrenfest Urn model. This means that it is necessary to consider an aggregate process $X_n = a[x]_n$, in which all configurations corresponding to the same number of balls in the first urn are lumped together. This aggregate process is reversible, as it is derived from a reversible Markov Chain (every time homogenous, two state Markov Chain is reversible, [Whittle, 1986]). It is also a Markov Chain because the aggregation procedure corresponds to a maximal invariant under permutations of balls, which preserve the statistics of the process, [Whittle, 1986, theorem 3.7.2]. This can be seen from the fact, that permuting two arbitrary balls in the same urn does not affect the probability distribution of the process and the lumping procedure described above establishes equivalence classes of states which have the same probability. Therefore summing equation (4) over all configurations $x$ corresponding to the same number of balls in the first urn $a[x]$ one obtains a probability distribution of the generalised Ehrenfest Urn model and hence that of SDS:

$$\pi(n) = \binom{N}{n} \pi_1^n \pi_2^{N-n} \tag{5}$$

which is a binomial distribution.

Equation (5) describes the steady state probability distribution of the whole ensemble of agents used in the Stochastic Diffusion Search. It describes probabilities of finding agents in different configurations of states, which implies different possible distributions of resources by SDS. We can characterise the resource allocation of SDS in the steady state by computing the expected distribution of agents. The deviations from the expected distribution can be characterised in terms of the standard deviation of the ensemble probability distribution.

From equation (5) it follows immediately that the expected number of active agents in the equilibrium will be

$$E[n] = N\pi_1 \tag{6}$$

In fact the most likely state, n, given by the maximum of the binomial distribution will be an integer number fulfilling following inequalities, [Johnson, Kotz, 1969],

$$(N+1)\pi_1 - 1 \leq n \leq (N+1)\pi_1$$

This implies that for sufficiently large $N$ the expected number of active agents is a good characterisation of their actual most likely number.

Similarly the standard deviation defined for binomial distribution as

$$\sigma = \sqrt{N\pi_1\pi_2} \tag{7}$$

will be used as a measure of variability of the number of active agents around its steady state.

In fact E[$n$] is not sensu-stricto an equilibrium of the system. From strong ergodicity it follows that eventually all possible configurations of agents will be encountered provided that one would wait sufficiently long. However, as in the case of the system with two containers with a gas, although all states are possible, nevertheless some of them will appear extremely rarely (e.g. state in which all the particles of the gas would concentrate in one container only). In fact, the system will spend most of the time fluctuating around the expected state, which thus can be considered as quasi-equilibrium.

The above discussion closely parallels the reasoning motivating Ehrenfests in using their model for a discussion of an apparent contradiction between the reversible laws of micro-dynamics of single particles and irreversibility of thermodynamic quantities.

The next section will illustrate some of these points with a numerical example and will characterise the resource allocation of SDS in terms of the search space and the properties of the best solution.

## *4. Characterisation of the resource allocation.*

The population of active agents converging to the best-fit solution is very robust. The convergence rate and the best-fit population size clearly differentiate SDS from a pure parallel random search [Nasuto, Bishop, 2002].

Below we will characterise the behaviour of the average number of active agents as a function of the parameter $p^-$ characterising the possibility of false negative response of agents for the best-fit solution. Figure 2 illustrates this relationship for the value of $p_m=0.001$ and $N=1000$ agents and Figure 3 shows the two dimensional plot of the average number of active agents as a function of both parameters $p_m$ and $p^-$.

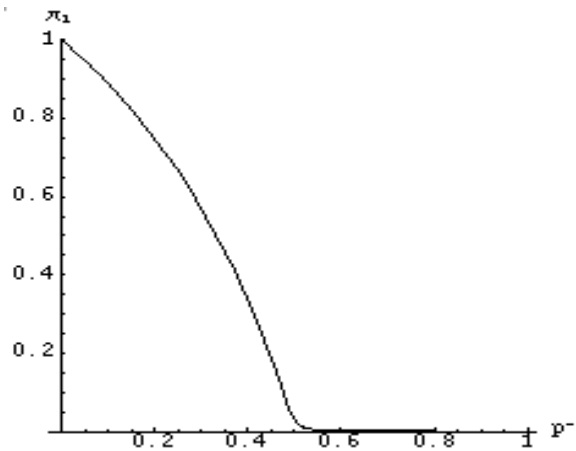

**Figure 2.** The normalised average number of active agents in SDS as a function of the parameter $p^-$. Plot obtained for $N=1000$ and $p_m=0.001$.

Figure 2 implies that the number of active agents decreases nonlinearly with an increase of the false negative parameter $p^-$ and reaches very small values around $p^-=0.5$. Thus, two regions of different characteristics of the resource allocation by SDS can be inferred from Figure 2. The first one is for $p^-<0.5$, where the cluster of active agents constitutes a significant part of the total amount of resources, and the second one is for $p^->0.5$, where the amount of active agents is orders of magnitude smaller.

From the fact that the number of agents in SDS is always finite it follows that for a given total number of agents there exists such value of $p^-$ that, above it, the actual number of active agents is almost always zero. This seems to confirm an estimate obtained in [Grech-Cini, 1995]. However, $\pi_1$ as a function of $p^-$ is everywhere positive in [0,1). It follows that for any $p^->0$ there exists a finite number of agents $N$ such, that $\lfloor N\pi_1 \rfloor > 0$, where $\lfloor x \rfloor$ denotes the greatest integer smaller than $x$.

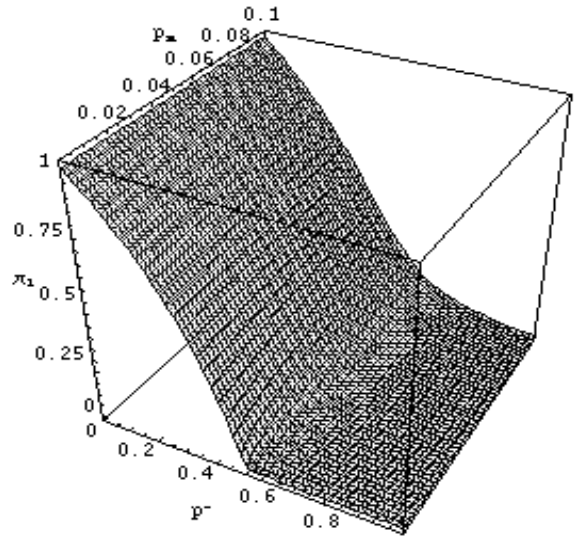

**Figure 3.** The normalised mean number of active agents as a function of both the false negative parameter $p^-$ and the probability of hit at the best instantiation $p_m$; plotted for $N$=1000.

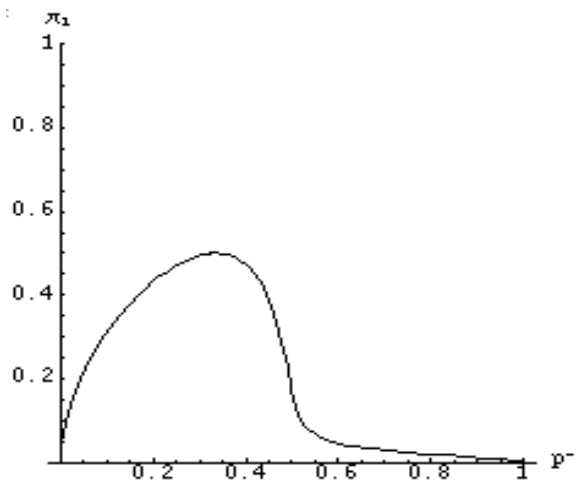

**Figure 4.** The rescaled standard deviation of the number of active agents calculated from the model for $N$=1000 agents and $p_m$=0.001; the scaling factor is $\alpha = N^{\frac{-1}{2}} \approx 0.0316$.

Figure 3 shows the normalised mean number of active agents as a function of both the false negative parameter $p^-$ and the probability of locating the best instantiation $p_m$. From the inspection of this figure it follows that changing $p_m$ does not significantly alter the dependence of the mean number of active agents on the parameter $p^-$. The change, resulting from an increase of $p_m$, can be summarised as a smoothing out of the boundary between two regions of behaviour of SDS clearly visible in Figure 2.

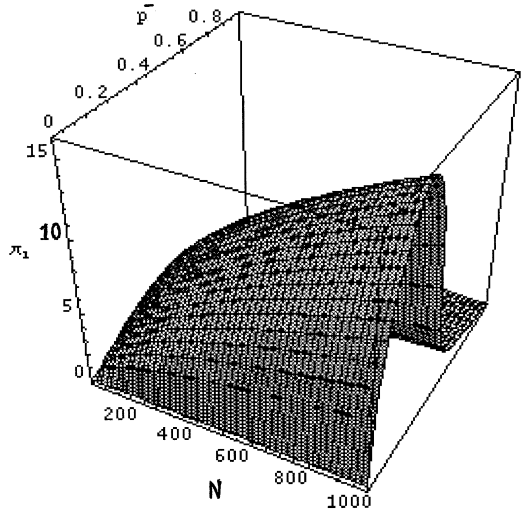

**Figure 5** The standard deviation of the number of active agents as a function of the total number of agents $N$ in SDS and of false negative probability $p^-$; plotted for $p_m=0.001$.

Similarly, it is possible to investigate, using the equations (3) and (7), dependence of the standard deviation on parameters of SDS. Figure 4 illustrates the behaviour of the standard deviation as a function of $p^-$, for $p_m=0.001$ and $N=1000$ agents and Figure 5 shows the 3D plot of the standard deviation as a function of $p^-$ and $N$.

From figure 4 one can deduce that standard deviation is also a non-linear function of $p^-$, first rapidly increasing with $p^-$ increasing from 0 to around 0.4, where the normalised standard deviation is largest, and then rapidly decreasing for $p^-$ increasing from 0.4 to 0.6. When $p^-$ increases further from 0.6 to 1 the normalised standard deviation decreases more steadily to 0. Figure 4 corresponds in fact to a cross-section of Figure 5 along the line of a fixed number of agents.

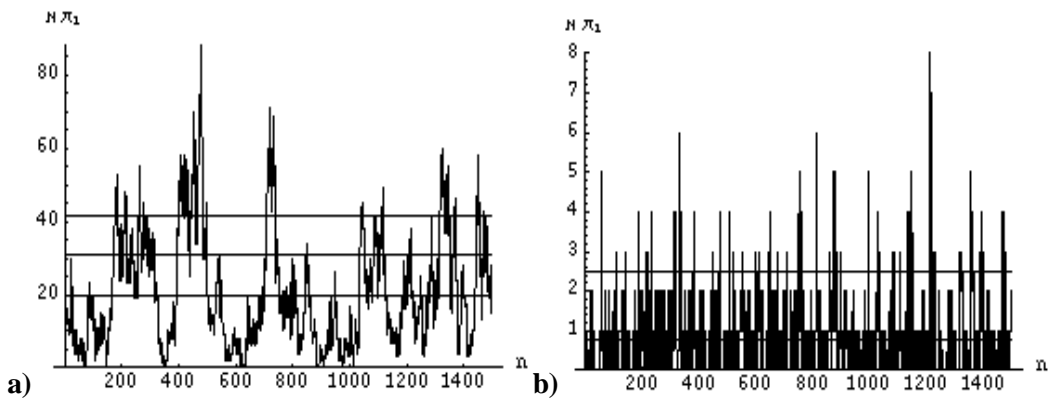

**Figure 6** Evolution of the number of active agent in SDS with $N=1000$ agents and $p_m=0.001$. The false negative parameter $p^-$ is 0.5 (left panel) and 0.7 (right panel). The straight lines correspond to the average activity predicted by the model surrounded by the +/-2 standard deviations band.

*5. Simulations.*

In this section we will evaluate the predictions of the Interacting Markov Chains SDS model by applying SDS to perform a string search, a simple yet nontrivial problem with very important applications in bioinformatics [Gusfield, 1997], information retrieval and extensions to other data structures [van Leeuven, 1990]. The task will be to locate the best instantiation of the given string template in the longer text. The agents will perform tests by comparing single characters in the template and the hypothesised solution and during the diffusion phase they will exchange information about potential location of the solution in the text. This set-up allows us to control precisely the parameters characterising the search space and the best-fit solution, hence it constitutes an appropriate test-bed for the evaluation of our model.

We run a number of simulations in order to compare reliably the theoretical estimates characterising quasi equilibrium with the actual behaviour of the system. The simulations reported here were run with $N$=1000 agents, $p_m$=0.001 and $p^-$ assuming values of 0.1, 0.2, 0.5 and 0.7 respectively. For calculating the estimates of the expected number of active agents and its standard deviation SDS was run for 2000 iterations. In all cases the first 500 samples were discarded as a burn-in period. This method was suggested in [Gilks et al, 1996] in order to avoid the bias caused by taking into account samples from the evolution when the process is far from the steady state.

| Average activity | Standard deviation |
|---|---|
| 889.02 | 10.06 |
| 749.61 | 15.22 |
| 20.57 | 15.48 |
| 0.81 | 1.09 |

**Table 1.** Average activities and standard deviations estimated from the 1500 iterations of SDS. $N$=1000, $p_m$=0.001 and $p^-$ changes from 0.1, 0.2, 0.5 to 0.7 respectively (top to bottom).

However, the size of the burn-in period is, in general, a difficult issue because it is related to the estimation of the convergence of a given Markov Chain to its steady state probability distribution. In the Markov Chain Monte Carlo practice, the number of iterations needed to obtain reliable estimates of statistics is often of the order of tens of thousands and the burn-in periods lengths can also be considerably long [Gilks et al, 1996]. The particular value of the burn-in period, used in these simulations, was chosen on the basis of visual inspection, as a heuristic remedy against the bias of estimates.

The results of the simulations are summarised in Table 1 and Table 2 and in Figure 6 and Figure 7.

| Average activity | Standard deviation |
|---|---|
| 888.06 | 9.97 |
| 750.1 | 13.69 |
| 30.65 | 5.45 |
| 0.75 | 0.86 |

**Table 2.** Average activities and standard deviations of SDS predicted by the model. $N=1000$, $p_m=0.001$ and $p^-$ changes from 0.1, 0.2, 0.5 to 0.7 respectively (top to bottom).

It follows that the model predicts very well the steady state behaviour of SDS. The biggest deviation between model prediction and empirical estimates seem to be for the value of $p^-=0.5$, i.e. in the case when agents have only 50% chance of successful testing of the best-fit solution. This may be because this value is located in the boundary of two regions of markedly different characteristic of resource allocation exhibited by SDS and small fluctuations in the number of agents in the cluster can have significant effects.

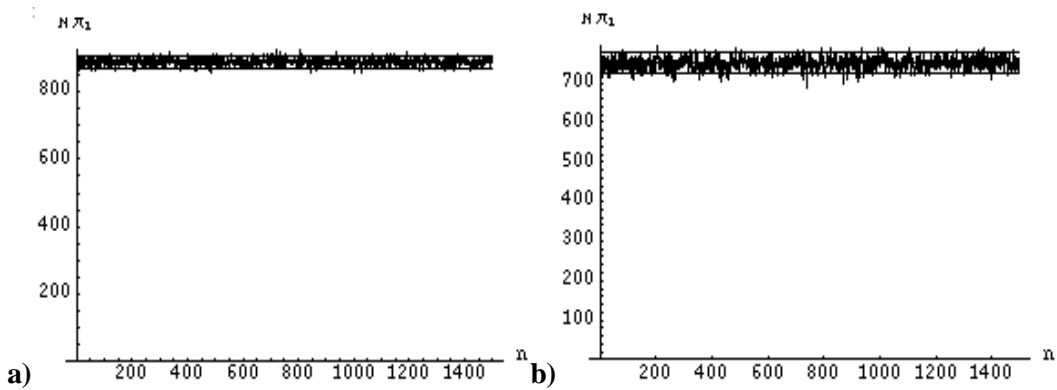

**Figure 7.** Evolution of the number of active agents in SDS with $N=1000$ agents and $p_m=0.001$. (a) The false negative parameter $p^-$ is 0.1; (b) $p^-=0.2$. The straight lines correspond to the average activity predicted by the model, surrounded by the +/-2 standard deviations band.

It appears that the expected number of active agents and two standard deviations, as calculated from the model, correspond well to the empirical parameters used in [Bishop, 1989] to define statistical stability of SDS and thus defining the halting criterion used in that work. Therefore our model provides a firm ground for the calculation of these parameters and for the characterisation of their dependence on the parameters of SDS and the search space.

*6. Conclusions.*

We have discussed in detail the resource allocation characteristics of SDS. It appears that SDS can rapidly and reliably allocate a large amount of agents to the best-fit pattern found in the search space, while the remaining agents continually exploring the search space for potentially better solutions. This property underlies SDS success in dealing with dynamic problems such as lip tracking, where the desired solution changes continually its position in the search space [Grech-Cini, 1995]. The analysis herein revealed that SDS is very sensitive to the quality of the match of the hypothesised solution. This property combined with the limited resource induced competition is responsible for the robustness of the algorithm to the presence of sub-optimal solutions, or noise in the search space [Nasuto et al, 1998]. Two key properties of SDS - the exchange of information between the agents and their relative computational simplicity illustrate the fact that the communication-based systems are as powerful as systems based on distributed computation, a thesis discussed in [Mikhailov, 1993]. Stochastic Diffusion Search can be easily extended to incorporate more sophisticated or heterogeneous agents, different forms of information exchange, including local communication, voting, delays etc. This makes it an appealing tool for both solving computationally demanding problems and simulating complex systems. The latter often are composed of many interacting subcomponents and it may be easier to define the operation of subcomponents and their mutual interactions, than find emergent properties of the system as a whole. We envision that SDS can constitute a generic tool for simulating such systems, e.g. in models of ecological systems, epidemics or agent based economic models.

SDS belongs to a broad family of heuristic algorithms, which utilise some form of feedback in their operation. Neural networks, a well-known class of such algorithms are often using some form of negative feedback during 'learning', i.e. during process of incremental change of their parameters guided by the use of disparity between desired and current response. In contrast, SDS, together with Genetic Algorithms and Ant Colonies, belongs to category of algorithms utilising positive feedback [Dorigo, 1994]. Positive feedback is responsible in SDS for its rapid convergence and, together with limited resources, for the non-linear dependence of the resource allocation on the similarity between the best-fit solution and the template. The resource allocation analysis presented in this paper, complements previous results on the global convergence and time complexity [Nasuto, Bishop, 1999; Nasuto, et al, 1998], setting a solid foundation for our understanding of its behaviour.

The resource allocation management of SDS was characterised by finding the steady state probability distribution of the (long-term behaviour) equivalent generalised Ehrenfest Urn model and observing that the actual distribution of agents is well described by the first two moments of this distribution.

It thus appeared that the resource allocation by SDS corresponds to the quasi-equilibrium state of the generalised Ehrenfest Urn model. This quasi equilibrium is well characterised by the expected number of active agents and the stability region, in which SDS will fluctuate - by two standard deviation bands around expected number of active agents. Finding explicit expressions for these quantities made it possible to characterise their dependence on the parameters of the search - a total number of agents, probability of false negative and a probability of randomly locating the best instantiation in a random draw. This analysis allowed us to understand how the resource allocation depends on the search conditions. This analysis also provides a theoretical background for the halting criterion introduced in [Bishop, 1989].

Apart of the use in analysis of SDS, the concept of interacting Markov Chains is interesting in its own right. Many computational models based on distributed computation or agent systems utilise a notion of semi autonomous computational entities, or agents. Such models are utilised in machine learning and computer science but also in models of economic behaviour [Young et al, 2000]. At some level of abstraction the evolution can be described in terms of Markov Chains. Interacting Markov Chains correspond to the situation when agents are not mutually independent but interact with each other in some way. This is often the case when agents try collectively to perform some task or achieve a desired goal. The basic formulation of interacting Markov Chains is flexible and can be extended to account for other types of interactions including heterogenous agents, delays in communication, rational as well as irrational agents, etc. Thus, it offers a very promising conceptual framework for analysing such systems.

*Figure Legends*

**Figure 1.** Schematic of Stochastic Diffusion Search

**Figure 2.** The normalised average number of active agents in SDS as a function of the parameter $p^-$. Plot obtained for $N$=1000 and $p_m$=0.001.

**Figure 3.** The normalised mean number of active agents as a function of both the false negative parameter $p^-$ and the probability of hit at the best instantiation $p_m$; plotted for $N$=1000.

**Figure 4.** The rescaled standard deviation of the number of active agents calculated from the model for $N$=1000 agents and $p_m$=0.001; the scaling factor is $\alpha = N^{\frac{-1}{2}} \approx 0.0316$.

**Figure 5** The standard deviation of the number of active agents as a function of the total number of agents $N$ in SDS and of false negative probability $p^-$; plotted for $p_m$=0.001.

**Figure 6** Evolution of the number of active agent in SDS with $N$=1000 agents and $p_m$=0.001. The false negative parameter $p^-$ is 0.5 (left panel) and 0.7 (right panel). The straight lines correspond to the average activity predicted by the model surrounded by the +/-2 standard deviations band.

**Figure 7.** Evolution of the number of active agents in SDS with $N$=1000 agents and $p_m$=0.001. (a) The false negative parameter $p^-$ is 0.1; (b) $p^-$=0.2. The straight lines correspond to the average activity predicted by the model, surrounded by the +/-2 standard deviations band.

*Table legends*

**Table 1** Average activities and standard deviations estimated from the 1500 iterations of SDS. $N$=1000, $p_m$=0.001 and $p^-$ changes from 0.1, 0.2, 0.5 to 0.7 respectively (top to bottom).

**Table 2** Average activities and standard deviations of SDS predicted by the model. $N$=1000, $p_m$=0.001 and $p^-$ changes from 0.1, 0.2, 0.5 to 0.7 respectively (top to bottom).

*Appendix*

*Proof of Proposition 1.* The proposition will be first proven for $p^->0$. Recall that

$$P_n = \begin{matrix} & a & n \\ a & \\ n \end{matrix} \begin{bmatrix} 1-p^- & p^- \\ p_1^n & 1-p_1^n \end{bmatrix}, \tag{A.1}$$

where

$$p_1^n = \tfrac{m}{N}(1-p^-) + (1-\tfrac{m}{N})p_m(1-p^-). \tag{A.2}$$

Define after (Seneta, 1973):

$$\lambda(P_n) \stackrel{d}{=} \max_j \left\{ \min_i p_{ij} \right\} =$$
$$= \max\{\min\{1-p^-, p_1\}, \min\{p^-, 1-p_1\}\}$$

Rearranging (A.2) leads to

$$p_1^n = (1-p^-)\big((1-p_m)x + p_m\big),$$

where *x* denotes the average activity in SDS. It is clear that $p_1^n$ is a linear function of *x* and as $0<x<1$, it follows that

$$p_1^n \in \left[ p_m(1-p^-), 1-p^- \right]$$

so

$$p_1^n \leq 1-p^- \;\Rightarrow\; \min(p_1^n, 1-p^-) = p_1^n$$
$$\Rightarrow\; \min(p^-, 1-p_1^n) = p^-$$

and therefore

$$\lambda(P_n) = \max(p_1^n, p^-).$$

Consider a series $\sum_{i=1}^{\infty} \lambda(P_n)$. This series is divergent because

$$(\forall n \geq 0)(\lambda(P_n) \geq p^-) \Rightarrow \sum \lambda(P_n) \geq \sum p^-$$

The last series diverges and therefore the weak ergodicity of $\{P_n\}$ follows from theorem 4.8 in [Seneta, 1973].

When $p^- = 0$ the same argument applies with the lower bound series taking the form $\sum_{i=1}^{\infty} p_m$. This is because from (A.2) it follows that $p_m$ is a lower bound for $p_1^n$. $\square$

*Proof of Proposition 2.* The above assertion can be proven by formulating the problem in geometric terms and showing that appropriately defined mapping has a fixed point. Consider the subset $K$ of a space $R^6$ defined as

$$K = X \times M_2^{p^-},$$
$$X = \{(p, 1-p) \mid 0 \leq p \leq 1\},$$
$$M_2^{p^-} = \left\{ \begin{bmatrix} 1-p^- & p^- \\ p_{21} & 1-p_{21} \end{bmatrix} \middle| \begin{array}{l} p_{21} = p(1-p^-) + (1-p)p_m(1-p^-) \\ = (1-p^-)(1-p_m)p + (1-p^-)p_m, \\ 0 \leq p_m, p^- \leq 1, (p, 1-p) \in X \end{array} \right\}$$

$X$ is a set of two dimensional probability vectors $(p, 1-p)$ and $M_2^{p^-}$ is the set of two dimensional stochastic matrices with fixed first row and components of the second row being linear functions of $p$. All points of $K$ can be attained by varying the parameter $p$ so by definition $K$ is a one dimensional subset of $R^6$. As $K$ can be thought of as a Cartesian product of one dimensional closed intervals, it follows that $K$ is convex and compact as a finite Cartesian product of convex, compact sets.

Define a norm in $R^6$

$$\| \| = \| \|_X + \| \|_M$$

where $\| \|_X$ and $\| \|_M$ are $l_1$ norm in $R^2$ and $l_1$ induced matrix norm in $M_2^{p^-}$ respectively.

Define the mapping

$$S: K \to K,$$
$$S(p, P) = (pP, P_{pP}),$$

where

$$p = (p, 1-p),$$
$$P = \begin{bmatrix} 1-p^- & p^- \\ p_{21} & 1-p_{21} \end{bmatrix},$$
$$P_{pP} = \begin{bmatrix} 1-p^- & p^- \\ p_{21}^* & 1-p_{21}^* \end{bmatrix},$$

and

$$pP = (p^*, 1-p^*) = (p(1-p^-) + (1-p)p_{21}, pp^- + (1-p)(1-p_{21})),$$
$$p_{21} = p(1-p^-) + (1-p)(1-p^-)p_m = (1-p_m)(1-p^-)p + p_m(1-p^-), \quad (A.3)$$

It follows that

$$\begin{aligned} p^* &= p(1-p^-) + (1-p)((1-p_m)(1-p^-)p + p_m(1-p^-)) = \\ &= (1-p^-)[p + (1-p_m)(1-p)p + (1-p)p_m] \\ &= (1-p_m)(1-p^-)p(2-p) + p_m(1-p^-) \\ p_{21}^* &= (1-p_m)(1-p^-)p^* + p_m(1-p^-) \end{aligned} \quad (A.4)$$

*S* acts on both components of a point from *K* returning probability distribution and stochastic matrix obtained as a result of one step evolution of nonhomogenous Markov Chain corresponding to the one step evolution of an agent.

It is possible to prove that *S* is continuous in *K*. In order to show this, choose an $\varepsilon > 0$ and fix an arbitrary point in *K*, (*q*,*Q*) and chose another point (*p*,*P*), such that

$$\|(q, Q) - (p, P)\| \leq \delta.$$

Note that

$$\|q - p\|_X = |q - p| + |1 - q - 1 + p| =$$
$$= |q - p| + |-q + p| = 2|q - p|$$

and that using (A.3)

$$\|Q - P\|_M = \left\|\begin{bmatrix} 1 - p^- & p^- \\ q_{21} & 1 - q_{21} \end{bmatrix} - \begin{bmatrix} 1 - p^- & p^- \\ p_{21} & 1 - p_{21} \end{bmatrix}\right\| =$$
$$= \left\|\begin{bmatrix} 0 & 0 \\ q_{21} - p_{21} & p_{21} - q_{21} \end{bmatrix}\right\| = |q_{21} - p_{21}| =$$
$$= (1 - p^-)(1 - p_m)|q - p| = \tfrac{1}{2}(1 - p^-)(1 - p_m)\|q - p\|_X$$

Together the above give

$$\|(q, Q) - (p, P)\| = \|q - p\|_X + \|Q - P\|_M = \left[1 + \tfrac{1}{2}(1 - p^-)(1 - p_m)\right]\|q - p\|_X .$$

Now

$$\|S(q, Q) - S(p, P)\| = \|(qQ, Q_{qQ}) - (pP, P_{pP})\| = \|(qQ - pP, Q_{qQ} - P_{pP})\| =$$
$$= \|qQ - pP\|_X + \|Q_{qQ} - P_{pP}\|_M$$

Both terms in the above equation will be considered separately.

$$\|qQ - pP\|_X = 2|q^* - p^*| =$$
$$= 2(1 - p_m)(1 - p^-)|q(2 - q) - p(2 - p)| = 2(1 - p_m)(1 - p^-)|2(q - p) - (q^2 - p^2)| =$$
$$= 2(1 - p_m)(1 - p^-)|2 - (q + p)||q - p| = (1 - p_m)(1 - p^-)|2 - (q + p)|\|q - p\|_X$$

Similarly

$$\|Q_{qQ} - P_{pP}\|_M = |q^*_{21} - p^*_{21}| = |(1 - p_m)(1 - p^-)(q^* - p^*)| =$$
$$= \tfrac{1}{2}(1 - p_m)^2(1 - p^-)^2|2 - (q + p)|\|q - p\|_X = (1 - p_m)(1 - p^-)|2 - (q + p)|\|Q - P\|_M$$

Finally it follows that

$$\|S(q, Q) - S(p, P)\| = (1 - p^-)(1 - p_m)|2 - (p + q)|\left[1 + \tfrac{1}{2}(1 - p^-)(1 - p_m)\right]\|q - p\|_X =$$
$$(1 - p^-)(1 - p_m)|2 - (p + q)|\|(q, Q) - (p, P)\|$$

Continuity of the operator S follows from the fact that for $\delta = \frac{\varepsilon}{2(1-p^-)(1-p_m)}$ one obtains

$$\|S(q,Q) - S(p,P)\| = (1-p^-)(1-p_m)|2-(p+q)|\|(q,Q)-(p,P)\| \leq$$
$$\leq 2(1-p^-)(1-p_m)\|(q,Q)-(p,P)\| \leq \varepsilon$$

Thus, by the Birkhoff-Kellogg-Schauder fixed point theorem, [Saaty, 1981], it follows that S has a fixed point in K. This property implies that the sequence $\{P_i\}$ of stochastic matrices is asymptotically stationary. □